\documentclass[10pt,twocolumn,letterpaper]{article}

\usepackage{cvpr}
\usepackage{times}
\usepackage{epsfig}
\usepackage{graphicx}
\usepackage{amsmath}
\usepackage{amssymb}

\usepackage{subfig}
\usepackage{multirow}

\usepackage[pagebackref=true,breaklinks=true,letterpaper=true,colorlinks,bookmarks=false]{hyperref}

\cvprfinalcopy % *** Uncomment this line for the final submission

\def\cvprPaperID{****} % *** Enter the CVPR Paper ID here

\ifcvprfinal\fi
\begin{document}

%%%%%%%%% TITLE
\title{Probing the State of the Art: \\ A Critical Look at Visual Representation Evaluation}

\author{Cinjon Resnick\\
NYU\\
{\tt\small cinjon@nyu.edu}
\and
Zeping Zhan \\
NYU \\
{\tt\small zz2332@nyu.edu}
\and
Joan Bruna \\
NYU \\
{\tt\small bruna@cims.nyu.edu}
}

\maketitle
%\thispagestyle{empty}

%%%%%%%%% ABSTRACT
\begin{abstract}
   Self-supervised research improved greatly over the past half decade, with much of the growth being driven by objectives that are hard to quantitatively compare. These techniques include colorization, cyclical consistency, and noise-contrastive estimation from image patches. Consequently, the field has settled on a handful of measurements that depend on linear probes to adjudicate which approaches are the best. Our first contribution is to show that this test is insufficient and that models which perform poorly (strongly) on linear classification can perform strongly (weakly) on more involved tasks like temporal activity localization. Our second contribution is to analyze the capabilities of five different representations. And our third contribution is a much needed new dataset for temporal activity localization.
\end{abstract}

%%%%%%%%% BODY TEXT
\section{Introduction}
There have been plenty of recent advances in self-supervised visual models \cite{hjelm2018learning,DBLP:journals/corr/abs-1807-03748,DBLP:journals/corr/abs-1905-09272}. These have all been built on the back of prior work \cite{DBLP:journals/corr/DoerschGE15,DBLP:journals/corr/ZhangIE16,DBLP:journals/corr/NorooziF16,zhang2016colorful,Deshpande2015LearningLA,DBLP:journals/corr/LarssonMS16} and are edging closer to the capabilities of supervised models in terms of image classification and detection, two tasks used in the literature to discriminate progress since at least 2017 \cite{DBLP:journals/corr/abs-1708-07860}. At that time, the SOTA was 68.6\% on ImageNet top-5 classification and 69.5\% on Pascal VOC Detection mAP, compared to 85.1\% and 74.2\% for the supervised models. Results today are now substantially better \cite{DBLP:journals/corr/abs-1901-09005}. For example, Pascal VOC is now at 75.8\% for supervised models and 73.0\% for self-supervised ones. Further, self-supervised models seem to actually be better than supervised models for 3D Scene Understanding and Visual Navigation \cite{DBLP:journals/corr/abs-1905-01235}.

\begin{figure}[t]
    \centering
    \includegraphics[width=\linewidth]{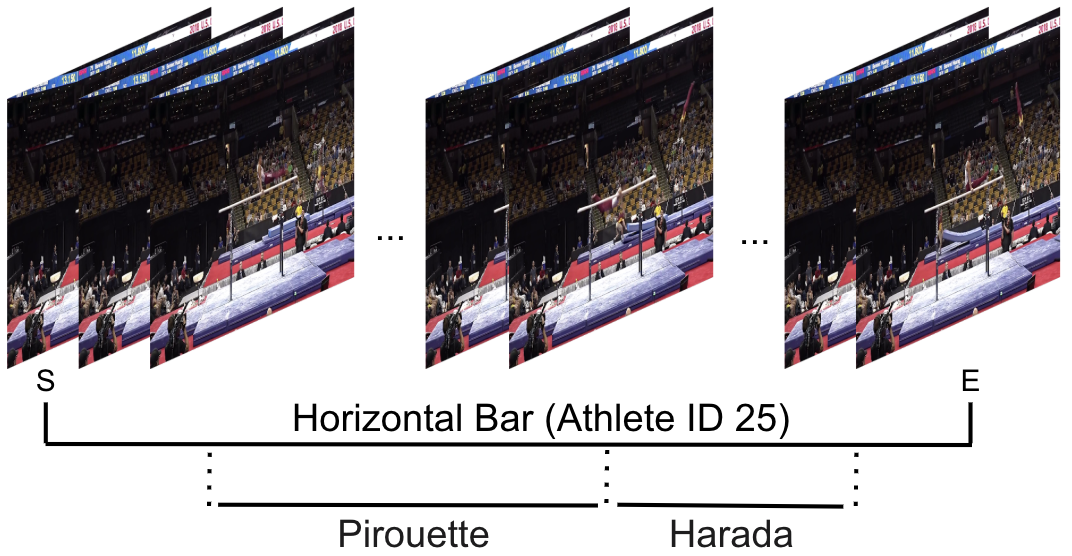}
    \caption{Graphic display of an example in the Gymnastics dataset. The thread is annotated with the apparatus, the athlete ID, and the start and end times. Each identified skill within that thread is also marked with start and end times.}
    \label{graphic:gymnastics}
\end{figure}

% We align with the statement that `A good representation should generalize to many tasks with limited supervision' (youtube.com/https://www.youtube.com/watch?v=ByM8WbX8Vqg 46:30 --> fueled the Facebook SSL Challenge
% - "No Full Fine-Tuning")

The standard methodology to assess self-supervised models is to extract the last layer of the trained representation and graft a linear classifier on top -- we refer to this as \emph{linear probing}. It is from this approach that we yield the numbers above. While the improved results themselves suggest the need for a new task, we also question whether linear probes are a suitable test regardless. They have been used to date as an easy way to compare learned representations from models because we otherwise have limited means to probe at the model itself. However, we claim that what the research is driving after is a strong representation space that is easily transferred into whatever shape a new task needs. This is the promise of self-supervised learning where the data is abundant, arguably even infinite. The literature to date measures progress by assessing the ability to linearly separate image categories, but is that really a suitable test? Our first contribution is that we show an experiment that suggests that linearly separating images is an insufficient axis for adjudicating progress (Sec.~\ref{sec:classification-results}). This experiment additionally suggests that the architecture chosen for the self-supervised models may have an inordinately positive bias on their transfer capability.

Consequently, we propose to focus on a %believe that the community should be asking
a further question that aims to better discriminate these models: how capable are these representations in handling tasks that require more than just a static image? This is important because it challenges if the model has really learned notions of semantics and general visual understanding that are hard to delineate from just images. Our hypothesis was that the supervised models will have an advantage across the board on this task over the status quo self-supervised models. Our results suggest that the latter are stronger than we suspected and that their relative ordering may not be correct.

{Our second contribution is to analyze in-depth the capabilities of five different representations - three self-supervised and two supervised.} We choose temporal activity localization for this challenge. In this task, the model's objective is to bound where in a video is there action taking place. A successful model needs to have understanding that goes beyond single images. Further, the task needs nonlinear machinery on top of the image representations. It is also a task where high quality annotated data is in short supply. Thus, we would expect that improvements in self-supervised techniques would be of great help.

Finally, {our third contribution is that we debut a new dataset for temporal activity localization - Gymnastics}. Currently, Gymnastics is comparable to the popular Thumos14 dataset \cite{THUMOS14}, but is larger and has been designed in a more natural way and with temporal activity localization in mind. Most notably, it contains frequent instances of multi-class scenes, which are common in real-world action localization problems but do not appear in Thumos14. We use this dataset to constrast the results we see on Thumos14 on a sufficiently different distribution.

Overall, our paper should be considered a diagnostic on the state of the art, and our experiments hone in on areas that we would like to understand better or find troubling.

\section{Related Work}
The Visual Task Adaptation Benchmark \cite{zhai2019visual} was recently introduced. It is a suite of 19 tasks that are meant to cover a broad spectrum of domains and semantics. The tasks are grouped into three categories according to the type of image in the task - Natural, Specialized, and Structured. While some of them are quite involved, they all are inherently classification tasks on top of a single 2D or 3D image. In contrast, our chosen task of temporal localization cannot be solved with a single image and requires understanding of action over time. Consequently, it is a stronger test of how generalizable across tasks are the representations.

Another recent benchmark analyzing visual representations was the FAIR Self-Supervision Benchmark \cite{DBLP:journals/corr/abs-1905-01235}, which was accompanied by an associated suite of tools to evaluate the quality of the representations. The principles behind their chosen tasks were that a good representation (1) transfers to many different tasks and (2) transfers with limited supervision and limited fine-tuning. And the tasks themselves were image classification, object detection, visual navigation, and surface normal estimation. These tasks can be quite challenging and involve heavier machinery than just classification using linear probes, which we agree with and believe to be important for the field. Our contribution is that we ask how these representations do on tasks that require understanding of video.

There are many works that explore temporal activity localization \cite{DBLP:journals/corr/ShouCZMC17,DBLP:journals/corr/ShouWC16,DBLP:journals/corr/abs-1804-07667,DBLP:journals/corr/ZhaoXWWLT17,DBLP:journals/corr/GaoYN17,DBLP:journals/corr/GaoYSCN17,DBLP:journals/corr/XuDS17}. We did not try to build a better approach than these other works, but rather to use the task as machinery to better understand the differences between representations. Towards that end, we choose to build upon Boundary Sensitive Networks \cite{DBLP:journals/corr/abs-1806-02964} (\textbf{BSN}) as it is near state of the art on this task and additionally includes both an open source repository and responsive authors.

\section{Setup}

We performed a comparison of representations over five different models. For each of these models, we assessed how capable they were at classification on CIFAR-10 and CIFAR-100 and at temporal activity localization on Thumos14 and Gymnastics.

\subsection{The Models}
\label{sec:models}

The models can be broken up into two categories - supervised and self-supervised. We briefly describe them here and why we chose them.

Temporal Segment Networks \cite{DBLP:journals/corr/WangXW0LTG16} (\textbf{TSN}) was chosen because it is a popular supervised model for video recognition in addition to it being the features used in the original BSN paper. We were able to reproduce their results on Thumos14. For the Gymnastics results, we used the same procedure\cite{mmaction2019} that they did for yielding representations as the concatenation of RGB and Flow features. The training set for the TSN models was the UCF-101 dataset \cite{DBLP:journals/corr/abs-1212-0402}, which contains trimmed videos for action recognition. Note that we did not use TSN in the CIFAR experiments because there is no valid concept of optical flow in static images.

Residual Networks ({\bf ResNet}) \cite{DBLP:journals/corr/HeZRS15} was chosen because it is a popular supervised model for images. This was important to include because we also have self-supervised image models in our set of comparisons and because we otherwise would not have a supervised model in our CIFAR experiments. We use ResNet50 from the Torchvision repository \cite{Marcel:2010:TMP:1873951.1874254}, which yields $2048$ dimensional features on RGB frames regardless of dataset.

Augmented Multiscale DIM \cite{hjelm2018learning} (\textbf{AMDIM}) was chosen because it was at or near state-of-the-art for self-supervised image models when we began this research. The representations were taken over RGB frames and were the concatenation of each of the last layer outputs. All together, this produced a per-frame representation of dimension $192000$ regardless of dataset. It was trained using the Pytorch-Lightning framework \cite{Falcon2019} on ImageNet \cite{imagenet_cvpr09}.

Cycle Consistency of Time \cite{CVPR2019_CycleTime} (\textbf{TimeCycle}) was chosen because it is a recent self-supervised model for vision that exhibits a novel and capable approach for learning the semantics of video and tracking in particular. It does this by employing the cycle-consistency of sequential patches in a video. We used the model distributed by the authors, which was trained on VLOG \cite{Fouhey18}. The strength of this model depends on the size of the image crop, however the representations can be very large. For half of our models - those with a nonlinear conversion (see Sec.~\ref{sec:tal-procedure}) - we use a crop of $(256, 256)$. This produces a representation of size $933888$, which was too large for the other half of our models that do not convert the representation. In these cases, we reduced the crop to $(128, 128)$ with an associated representation of size $237568$. For the CIFAR experiments, we further reduced this to the size of the image, $(32, 32)$, which produced an $8192$ sized representation.

Video Correspondence Flow \cite{Lai19} (\textbf{CorrFlow}) was chosen because it is a self-supervised model for video that attempts to solve a similar problem to TimeCycle. It achieves stronger numbers by employing a common technique in the literature - colorization \cite{DBLP:journals/corr/abs-1806-09594}. Comparing this and TimeCycle on an unrelated task would potentially shed more light on the differences between their representations. We used the model distributed by the authors, which was trained on Kinetics \cite{DBLP:journals/corr/KayCSZHVVGBNSZ17} and has representation size $225280$ for our localization tasks and $4096$ for CIFAR.

\subsection{The Datasets}
We use Thumos14 and Gymnastics to compare the models on the temporal localization task. For the more common classification task, we use CIFAR-10 and CIFAR-100.

{\bf CIFAR} \cite{Krizhevsky09learningmultiple} is split into CIFAR-10 and CIFAR-100. The former has 6000 examples of each of 10 classes and the latter has 600 examples of each of 100 classes, where each example is a $(32, 32)$ image.

{\bf Thumos14} is the most popular dataset for temporal activity localization. While Thumos14 is quite large, a smaller subset of temporally annotated and untrimmed videos are commonly used for this task. In total, it has 200 training videos (taken from the validation set) and 213 test videos (taken from the test set) covering twenty action classes.

% {\bf ActivityNet} is another common dataset for comparing performance on this task. We pre-process it with ffmpeg to make all of the videos a uniform 24 frames-per-second and $(320, 240)$ size. Additionally, because ActivityNet has many videos and our model representations can be very large, we train on a random 30\% of the data that is resampled each epoch.

\subsubsection{Gymnastics}

Thumos14 has historically been considered the most appropriate dataset for temporal activity localization. Compared to previous datasets such as ActivityNet \cite{conf/cvpr/HeilbronEGN15}, Thumos14 has more action instances per video and each video contains a larger portion of background activity. However, it itself is rather small with only 11.6 hours of training data spanning twenty classes (detailed in Table \ref{table:thumos-tiou5}). Further, the videos are taken from YouTube and sourced in a largely automated fashion. While there are extensive steps taken to ensure that the videos are on topic, they are still guided by engineering intuition rather than intentional inclusion and thus allow for errors. The clean-up and validation steps are then performed by laymen and thus also do not satisfy the ideal levels of data cleanliness.

This is further exacerbated in that it is also laymen who provide the temporal annotations. These can be quite dirty, especially on the boundaries where actions begin and end, which are the most difficult part of the task at hand. In addition, the videos often include moving cameras and hard cuts among scenes.

While data cleanliness is a high priority in any machine learning pipeline, this would not necessarily be the most pressing problem if the dataset was very large. However, while Thumos14 is a large dataset, the subset that is useful for temporal localization is not.

In comparison, the Gymnastics dataset started from a very different hypothesis than Thumos14 or ActivityNet. We aimed to solve real-world problems around human motion, starting with Gymnastics. This includes temporal localization but also other sought-after tasks such as error analysis and deduction awareness. Consequently, the dataset includes both full training sessions and competition performances. It is realistic in that it was filmed by gymnasts or coaches for their own library with cameras fixed and positioned to capture the entire session. The results were then expertly annotated and otherwise left untrimmed.

Currently, it contains 412 videos of men's gymnastics split into train (322), validation (44), and test (46), with train having 36.8 hours of data on five apparatuses: floor exercise, pommel horse, still rings, parallel bars, and horizontal bar\footnote{We did not include vault in this version because most of the data we have of that apparatus was non-stationary.}. Within that time, there are 14.1 hours of action split amongst 1940 threads. Each thread is annotated with the following information: the athlete ID, the start and end time within the larger video, the region of interest in the view, and the skills performed within that thread. An example thread would be an athlete performing a routine on horizontal bar. In total, there are 17618 annotated skills over 317 named classes for an average of 55 skills per class, each of which are also annotated with start and end within the thread. Examples of skills include Giants, Forward uprises, and Haradas. The mean duration of a skill is 2.27 seconds.
%Further, these are also annotated with point deduction on that remark.

Gymnastics also contains videos with concurrent threads, which consequently results in multi-class prediction tasks. This is a common feature of real videos of motion and which has historically been missing from academic datasets. ActivityNet has a very small number of these; Thumos14's annotations suggest that it has this feature but those are actually just mislabeled annotations of cut scenes.

Please see Table \ref{table:dataset-comparison} for a comparison of Gymnastics with both Thumos14 and ActivityNet.

\begin{table}
\begin{center}
{
\small
\begin{tabular}{|c|c|c|c|c|c|}
\hline
& Thumos14 & ActivityNet & Gymnastics \\ \hline
Total Videos & 200 & 9649 & 322 \\
Duration & 209.0 & 117.5 & 412.1 \\
Instances & 15.4 & 1.5 & 11.1 \\
Background & 64.0\% & 33.4\% & 27.3\% \\
Concurrent Labels & 0 & 0.018 & .957 \\ 
Source & YouTube & YouTube & Original Footage \\ \hline
\end{tabular}
}
\end{center}
\caption{Thumos14, ActivityNet, and Gymnastics training statistics. Besides total videos, all other columns are per-video. Duration is in seconds. For concurrent instances, the annotations in Thumos14 suggests that these exist, however the videos do not show them as such. ActivityNet lists its sources as `Online video sharing sites'. We interpret this to mean YouTube given the videos in the dataset.}
\label{table:dataset-comparison}
\end{table}

\section{Image Classification}
We show classification performance on CIFAR-10 and CIFAR-100 for three reasons. The first is that neither CorrFlow nor TimeCycle have public results on this task. It would be illuminating to ask how they compare to the rest of the literature given that they have only been trained on a self-supervised task over video. The second is that it strengthens our other results in that it verifies that the chosen representations for AMDIM and ResNet are satisfactory for approximately matching the known performance in the literature. The third reason is that we found a reproducible discrepancy for TimeCycle that suggests that linearly separating images is an insufficient test for adjudicating the success of self-supervised models.

\begin{table}[ht]
\begin{center}
{
\small
\begin{tabular}{|c|c|c|c|c|c|}
\hline
Pretrained? & Model & CIFAR-10 & CIFAR-100 \\ \hline
\multirow{4}*{Yes} & AMDIM & 87.7\% & 66.7\% \\
& CorrFlow & 59.7\% & 31.4\% \\
& ResNet & 90.5\% & 72.9\%\\
& TimeCycle & \textbf{67.2\%} & \textbf{38.9\%} \\ \hline
\multirow{4}*{No} & AMDIM & 55.4\% & 32.6\% \\
& CorrFlow & 52.0\% & 25.2\% \\ 
& ResNet & 37.6\% & 14.2\% \\ 
& TimeCycle & \textbf{81.3\%} & 
\textbf{61.8\%} \\ \hline
\end{tabular}
}
\end{center}
\caption{Classification performance on CIFAR-10 and CIFAR-100. Note the large boost in performance between pretrained and random TimeCycle. We do not use TSN because there is no notion of optical flow in this dataset and so consequently the results would be hard to meaningfully interpret.}
\label{table:cifar-classification}
\end{table}

\subsection{Procedure}
We train our models in two ways. The first way is by loading a pre-trained checkpoint (as specified in Sec.~\ref{sec:models}), freezing that model, and then training a linear classifier on top of the representations that model outputs. The second is to do the same but randomly initialize the network instead of loading the checkpoint. Table \ref{table:cifar-classification} delineates these two as, respectively, pretrained and not pretrained. The reason for this dual procedure is to ask how much the training procedure biases the ensuing representations over the bias inherent in the architecture itself.

\subsection{Results}
\label{sec:classification-results}

As seen in Table \ref{table:cifar-classification}, we attained strong results for AMDIM and ResNet on both CIFAR-10 and CIFAR-100 when using the trained checkpoint. We emphasize that these were both trained on ImageNet, and so these results show a nice capability of these models to generalize to CIFAR. Observe that both CorrFlow's and TimeCycle's results were much weaker.

We then duplicated this procedure but on a randomly initiated model that we then froze. As expected, AMDIM, ResNet, and CorrFlow all do worse on this task. TimeCycle, though, does a lot better on both CIFAR-10 and CIFAR-100. Its results are not much worse than trained AMDIM and ResNet and by far better than those models when randomly initialized. We were able to consistently reproduce this striking result with multiple seeds. This suggests first that the TimeCycle model architecture is quite capable of achieving high scores and, second, that its self-supervised learning objective biases the model towards a representation space that makes it difficult to linearly separate the images in CIFAR.

This raises an important question. Is the linear probe task actually the right one for discriminating self-supervised models? If we judged our progress on this marker, we would have considered both TimeCycle and CorrFlow to be poor models. However, they are not only strong at their learned tasks (both of which involve mask and pose propagation), but as will see in Section \ref{sec:temporal-localization-results}, they also produce representations that are quite capable for temporal localization. Arguably, they do a great job situating their representations in a way that allows for transfer to new tasks.

Further focusing on the randomly initialized results, we see that the self-supervised models all do better than ResNet on both CIFAR-10 and CIFAR-100. We cannot rule out the possibility that this is due to insufficient tuning, albeit we did try to give all of the models equal attention. This included a hyperparameter sweep as well as manual attention on more promising regions of the hyperparameter space. If we exclude that possibility though, this suggests that the architecture approaches taken in the self-supervised literature are already biasing the models towards good representations ignorant of the training objective itself.

% \begin{figure*}[ht]
%     \centering
%     \subfloat[Mean JS Divergence]{{\includegraphics[width=0.45\linewidth]{graphs/jsd_mean.png}}}
%     \hfill    
%     \subfloat[Similarity]{{\includegraphics[width=0.45\linewidth]{graphs/new_test_sim.png}}}
%     \caption{Two plots showing, respectively, the average JS divergence between any two model output distributions and the similarity of those models' representations, both over the test set. \cinjon{finish}}
%     \label{fig:sim-jsd-test}
% \end{figure*}

\begin{figure}[ht]
    \centering
    \includegraphics[width=\linewidth]{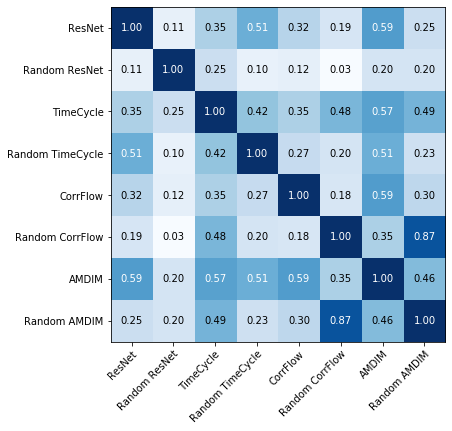}
    \caption{The CKA similarity between each pair of models' representations over the test set. Note that because the AMDIM features were too large for our machines, we used PCA to reduce their dimensionality from $192000$ to $8192$ before computing associated CKA similarities.}    
    \label{fig:sim-test}
\end{figure}

% \begin{figure}[h]
%     \centering
%     \includegraphics[width=\linewidth]{graphs/test_scatter.png}
%     \caption{A scatterplot of the Similarity to KL relationship in the CIFAR-10 test set. The train version is not shown here but is similar. We can see that there is a mostly inverse relationship between similarity and mean KL as expected. This is even before accounting for the pairs with very high KL - (RandomResNet, TimeCycle), (RandomResNet, ResNet), and (ResNet, TimeCycle).}
%     \label{fig:scatter-test}
% \end{figure}

Further comparing the representations, Fig.~\ref{fig:sim-test} shows the CKA similarity \cite{DBLP:journals/corr/abs-1905-00414} between each pair of models. It is a measure of how similar are their representation spaces. Recall from Table.~\ref{table:cifar-classification} that the best models were AMDIM, ResNet, and Random TimeCycle. For ResNet, we see that its representation space is relatively high with both of those models. However, we see that AMDIM has similar scores with Random AMDIM, CorrFlow, Random TimeCycle, and TimeCycle, suggesting that there is some sort of similarity occurring amongst the self-supervised representation spaces at large.

\section{Temporal Activity Localization}
\label{sec:temporal-activity-localization}

In this section, we detail our main task of temporal activity localization over both Thumos14 and Gymnastics.

\subsection{Procedure}
\label{sec:tal-procedure}
For each model, we load and freeze the pre-trained checkpoint. Then, for each dataset, we train a Boundary Sensitive Network on top of the model's representations.

We would like to be able to compare the representations from these different models on an entirely fair basis and address the question - \textit{for each model, how sufficient is the retained information inherent in its learned representation space towards the task of temporal activity localization?} However, we immediately run into a problem in that the BSN network was built for (frozen) representations from a Temporal Segment Network. This includes hyperparameter tuning but also model architecture and input size because the TSN dimensions are relatively small at 400 dimensions. The other networks have very different representation spaces and especially so when it comes to their dimension size. Tuning the BSN for each network would be prohibitively time-consuming as well as difficult to assess from the community's perspective. On the other hand, not tuning it at all would be unfair to the new representations.

We compromised by performing two approaches in order to make this as fair a comparison as we could:
\begin{itemize}
    \item For every self-supervised model $M$, we trained two versions: $M_{\text{NFC}}$ and $M_{\text{DFC}}$, where NFC stands for `No Feature Conversion' and DFC for `Do Feature Conversion'. The first, NFC, preserves the original feature representation length. The second, DFC, reduces the feature length to $400$ through a consistent nonlinear transformation, which we describe in Section \ref{par:nonlinear-transformation}.
    \item For every model $M$, we perform a hyperparameter sweep for both $M_{\text{NFC}}$ and $M_{\text{DFC}}$ on BSN.
\end{itemize}

With the $M_{\text{NFC}}$ model, we are asking the question - \textit{subject to limited capacity and potential optimization issues, does this representation space have sufficient information to inform the BSN?} 

With the $M_{\text{DFC}}$ model, we are asking the question - \textit{If we allow for better optimization, can this representation inform the BSN as well as TSN does?} 

We do not know a priori whether BSN paired with $M_{\text{NFC}}$ will even be capable of learning. We do expect though that $M_{\text{DFC}}$ will do better than $M_{\text{NFC}}$ because they provide $M$ with further (tuned) capacity. One can think of the $M_{\text{NFC}}$ representation as being a lower bound on the model's capability when fed to the BSN and the $M_{\text{DFC}}$ representation as being a transformation more suitable for the localization task and arguably with more capacity than using the TSN features for which the BSN was originally built.

There is another question that falls within these two bounds - \textit{can a linear transformation of the representation to 400 dimensions inform the BSN as well as TSN does?}. We think that this question is worthwhile but do not ask it as it falls between the two questions that we do query. This is because $M_{\text{NFC}}$ is equivalent to this approach up to when the linear transformation is performed. In $M_{\text{NFC}}$, the transformation is at the end, whereas this question would suggest putting it at the beginning. This could be done with, for example, a large fully connected network or principal component analysis.

\paragraph{Nonlinear Transformation}
\label{par:nonlinear-transformation}
Given the success of ResNet models as well as that all of our representations start with a ResNet base, we use a nonlinear ResNet-like transformation on top of CorrFlow, TimeCycle, and AMDIM to reduce the size of the representation to the same dimension as that of TSN. Explicitly, the transformation is a 7x7 convolutional network with stride of 2 and padding of 3, followed by batch norm and a ReLU, then two residual blocks\footnote{For further details, please see the original ResNet paper \cite{DBLP:journals/corr/HeZRS15}.} consisting of two blocks each with 64 channels and stride of 2. We then follow this transformation with a linear fully connected layer to yield a 400 dimensional representation.

\paragraph{Hyperparameter Sweep}
Our adaptation of BSN is a complex pipeline involving seven sequential steps:

\begin{enumerate}
    \item \textbf{Temporal Evaluation Module (TEM) training}: Train the TEM, which infers the starting, ending and action probabilities of each temporal location.
    \item \textbf{TEM evaluation}: Using validation results, evaluate which models from the hyperparameter sweep of TEM training were best. Continue with only these models.
    \item \textbf{TEM inference}: Run the TEM inference procedure over the entire dataset. This generates starting, ending, and action probabilities for use in the next phase.    
    \item \textbf{Proposal Generating Module}: With the inferred results, generate the proposals and features for the next set of modules.
    \item \textbf{Proposal Evaluation Module (PEM) training}: Train the PEM, yielding a confidence score for each proposal.
    \item \textbf{PEM evaluation}: Using validation results, evaluate which models from the hyperparameter sweep of PEM training were best. Continue with only these models.
    \item \textbf{PEM inference}: Run the PEM inference procedure over the test dataset, which generates refined proposals for each video.
    \item \textbf{Postprocessing \& Evaluation}: With the inferred results, post-process the results with soft non-maximum suppression and then output the final tIoU scores.
\end{enumerate}

In steps one and five, TEM training and PEM training respectively, we employ a hyperparameter sweep in order to elicit a more fair comparison among the representations. For step one, there are a total of 48 settings in the sweep. They range over: the initial learning rate, the step at which we reduce the learning rate, the gamma by which we reduce it, the L2 loss penalty, the weight decay penalty, and whether we use augmentation. For step five, there are eight settings in the sweep. These range over: the step at which we reduce the learning rate, the gamma by which we reduce it, the L2 loss penalty, and the weight decay penalty.

\section{Results}
\label{sec:temporal-localization-results}

Overall, our results suggest that supervised models are stronger than self-supervised models on the temporal localization task. However, there are nuances.

\begin{table}[h]
\begin{center}
{
\small
\begin{tabular}{|c|c|c|c|c|c|}
\hline
Reproduction & @50 & @100 & @200 & @500 & @1000 \\ \hline
TSN & \textbf{37.1} & \textbf{45.6} & \textbf{52.9} & \textbf{60.0} & \textbf{64.0} \\
CorrFlow DFC & 6.0 & 11.3 & 20.7 & 38.0 & 47.9 \\
CorrFlow NFC & 6.2 & 12.0 & 22.7 & 39.7 & 44.0 \\
TimeCycle DFC & 10.1 & 16.7 & 26.0 & 40.6 & 49.5 \\
ResNet DFC & 16.5 & 22.8 & 30.3 & 41.1 & 47.1 \\
ResNet NFC & 8.3 & 14.0 & 22.8 & 38.2 & 47.8 \\
AMDIM DFC & 12.4 & 18.6 & 27.7 & 40.3 & 44.4 \\ \hline
\end{tabular}
}
\end{center}
\caption{Comparison among representations on Thumos14 in terms of average recall at various proposal counts. Note that TimeCycle and AMDIM NFC never learned over the same amount of tuning sufficient for the other models.}
\label{table:thumos-proposal-avg-recall}
\end{table}

\begin{table}[h]
\centering
{
\small
\begin{tabular}{|c|c|c|c|c|c|c|}
\hline
Reproduction & @50 & @100 & @200 & @500 & @1000 \\ \hline
TSN & 25.2 & 29.8 & 34.9 & \textbf{38.7} & 38.7   \\
% TSN No Freeze & 14.3 & 17.6 & 21.7 & 24.8 & 24.9\\
% TSN DFC & 20.0 & 23.8 & 28.8 & 34.8 & 35.0\\
CorrFlow DFC & \textbf{35.1} & \textbf{36.3} & \textbf{37.7} & 37.8 & 37.8  \\
CorrFlow NFC & 21.8 & 26.3 & 30.2 & 30.2 & 30.2  \\
TimeCycle DFC & 29.4 & 31.9 & 33.9 & 33.9 & 33.9 \\
ResNet DFC & 26.7 & 31.1 & 33.3 & 37.6 & \textbf{46.9} \\
ResNet NFC & 20.8 & 25.5 & 29.5 & 34.5 & 45.6  \\
AMDIM DFC & 28.9 & 31.7 & 32.8 & 32.8 & 32.8   \\ \hline 
\end{tabular}
}
\caption{Comparison among representations on Gymnastics in terms of average recall at different numbers of proposals. Similarly to Thumos14, the AMDIM and TimeCycle NFC models were difficult to tune and did not learn anything coherent.}
\label{table:gymnastics-proposal-avg-recall}
\end{table}

\begin{table*}[h]
\begin{center}
{
\small
\begin{tabular}{|c|c|cc|c|cc|c|}
\hline
& \multicolumn{1}{c|}{TSN} 
& \multicolumn{2}{c|}{CorrFlow} 
& \multicolumn{1}{c|}{TimeCycle} 
& \multicolumn{2}{c|}{Resnet50}
& \multicolumn{1}{c}{AMDIM} 
\\
Category & Reproduction & DFC & NFC & DFC & DFC & NFC & DFC \\ \hline
Ambiguous & 87.8 & 85.8 & 78.8 & \textbf{87.9} & 86.7 & 83.4 & 76.8 \\
Baseball Pitch & \textbf{100.0} & 97.6 & 87.8 & 97.5 & \textbf{100.0} & \textbf{100.0} & 90.2 \\
Basketball Dunk & \textbf{76.4} & 69.5 & 64.5 & 69.0 & 61.9 & 64.3 & 63.3 \\
Billiards & \textbf{95.3} & 84.9 & 75.5 & 86.8 & 84.8 & 89.6 & 83.0 \\
Clean And Jerk & 91.8 & 86.7 & 85.7 & 91.8 & \textbf{92.9} & 92.8 & 91.8 \\
Cliff Diving & \textbf{99.1} & 97.2 & 90.3 & 98.6 & 96.8 & \textbf{99.1} & 85.7 \\
Cricket Bowling & \textbf{97.8} & 83.3 & 71.7 & 92.0 & 92.0 & 92.0 & 70.3 \\
Cricket Shot & \textbf{91.2} & 77.0 & 64.5 & 87.6 & 83.5 & 76.5 & 61.8 \\
Diving & \textbf{99.2} & 97.9 & 91.2 & 98.2 & 95.1 & 97.9 & 89.4 \\
Frisbee Catch & \textbf{100.0} & 95.8 & 93.7 & \textbf{100.0} & 97.9 & 95.8 & 91.7 \\
Golf Swing & \textbf{94.4} & \textbf{94.4} & 91.7 & \textbf{94.4} & \textbf{94.4} & \textbf{94.4} & 86.1 \\
HammerThrow & \textbf{87.6} & 71.9 & 81.2 & 84.7 & 81.4 & 72.7 & 78.9 \\
High Jump & \textbf{97.0} & 95.5 & \textbf{97.0} & \textbf{97.0} & \textbf{97.0} & 94.1 & 95.5 \\
Javelin Throw & \textbf{98.9} & 87.6 & 93.5 & 96.4 & 95.9 & 87.0 & 82.8 \\
Long Jump & \textbf{100.0} & 92.9 & 97.2 & 99.3 & 97.9 & 92.2 & 96.5 \\
Pole Vault & \textbf{98.5} & 87.2 & 90.9 & 96.2 & 95.5 & 90.0 & 92.5 \\
Shotput & \textbf{99.3} & 94.4 & 94.4 & \textbf{99.3} & 95.1 & 97.2 & 89.6 \\
Soccer Penalty & 97.9 & 93.7 & 93.7 & 93.7 & \textbf{100.0} & 97.9 & 87.5 \\
Tennis Swing & \textbf{94.3} & 80.8 & 73.0 & 88.6 & 86.5 & 85.8 & 76.6 \\
Throw Discus & \textbf{95.4} & 93.2 & 94.3 & 94.3 & 93.2 & 92.0 & 93.2 \\
Volleyball Spiking & \textbf{97.5} & 92.5 & 85.0 & \textbf{97.5} & 94.2 & 95.0 & 90.0 \\ \hline
\end{tabular}
}
\end{center}
\caption{Average recall on Thumos14 at tIoU of 0.5 for each class and model. As expected, TSN dominates, especially so in the following categories: Basketball Dunk, Billiards, Cricket Bowling, Cricket Shot, and Tennis Swing.}
\label{table:thumos-tiou5}
\end{table*}

\begin{figure*}[ht]
    \centering
    \subfloat[Thumos14 tIoU 100 Proposals]{{\includegraphics[width=0.4\linewidth]{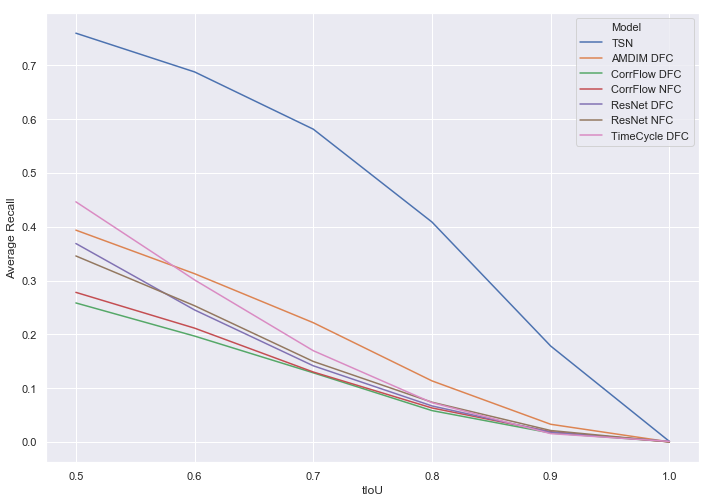}}}
    \qquad
    \subfloat[Thumos14 tIoU 1000 Proposals]{{\includegraphics[width=0.4\linewidth]{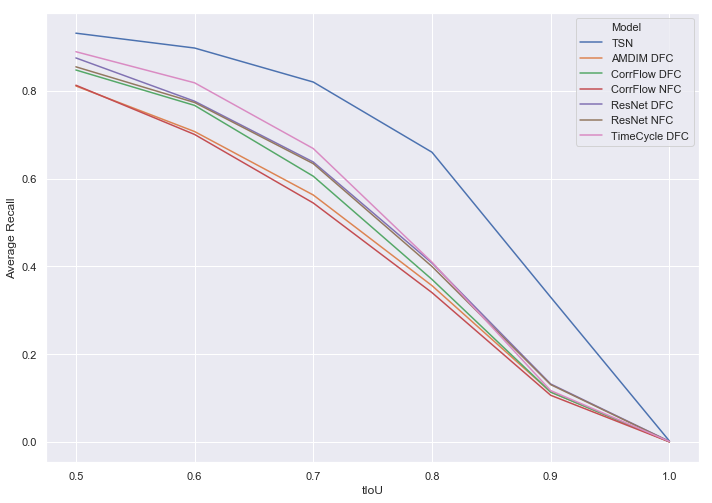}}}
    \qquad
    \subfloat[Gymnastics tIoU 100 Proposals]{{\includegraphics[width=0.4\linewidth]{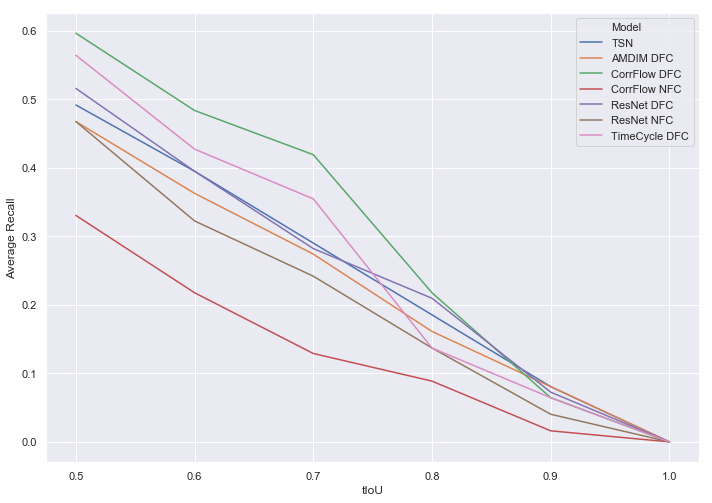}}}
    \qquad
    \subfloat[Gymnastics tIoU 1000 Proposals]{{\includegraphics[width=0.4\linewidth]{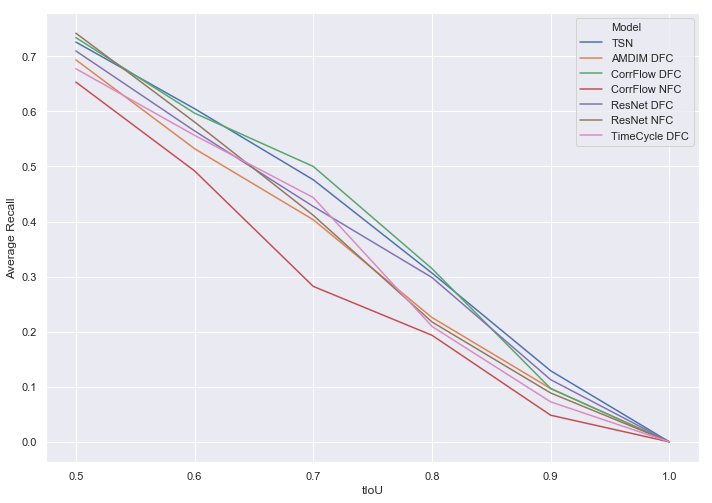}}}    
    \caption{Comparison of the models across different tIoU thresholds for each of 100 and 1000 proposals. Plots (a) and (b) show that TSN is better across all threshold levels, and TimeCycle, with a little bit of nonlinear tuning, can match that of ResNet. While plots (c) and (d) generally show how capable the self-supervised representations are when transferring to an unseen distribution, they particularly highlight CorrFlow's strong results at low numbers of proposals.}
    \label{fig:tiou-graphs}
\end{figure*}

We start with the Thumos14 results. Table \ref{table:thumos-proposal-avg-recall} shows that TSN does much better than any of the other models at every proposal count. This is further reinforced by Fig.~\ref{fig:tiou-graphs}~(a, b) where we see that TSN's average recall is far and away better for every tIoU threshold at both 100 and 1000 proposal counts. This is a common measure in temporal localization where the higher the better.

Diving deeper into the results, Table \ref{table:thumos-tiou5} shows the average recall percentage per category at tIoU of $0.5$. For all but two categories (Clean and Jerk and Soccer Penalty), we see that TSN has the highest score or is within $0.1$ of the highest score. In some cases (Basketball Dunk, Billiards, Cricket Bowling, and Tennis Swing), it is at least five percentage points better than the other models. And in the only categories where it is not the highest, it is ResNet, the other supervised model, that is the highest.

However, observe that the self-supervised models are still respectable in most categories. Aside from the aforementioned five particularly difficult categories, they collectively score well. And in many instances, TimeCycle does better than ResNet. Fig.~\ref{fig:tiou-graphs}~(a, b) even shows that TimeCycle DFC is on par or better than ResNet in terms of tIoU, suggesting that TimeCycle can be tuned to have as good a performance as a supervised model. This suggests that TimeCycle's representations are very amenable to transfer learning and better than what a linear probe task would suggest.

Recall further that TSN was trained on UCF-101. This dataset consists of actions that are similar to those in Thumos14, and the videos were sourced in a familiar manner (from YouTube) by the same research group. Arguably, it was trained on as close to the same distribution as one could without it being the actual same dataset.

On the other hand, the ResNet model we used was trained on ImageNet, which is a dissimilar dataset from Thumos14. We see that its numbers are on par with and even worse than all three of the self-supervised models. This suggests that the self-supervised representations are just as capable, if not more so in the case of TimeCycle, as a generically well-trained supervised model for this altogether different task.

To explore this further, we turn to our second dataset - Gymnastics. While we remain within the sports domain, our data is very different from UCF-101. This extends to both how it was sourced, who sourced it, and the video statistics (detailed in Table~\ref{table:dataset-comparison}). This should lessen the home court distribution advantage for TSN. Table \ref{table:gymnastics-proposal-avg-recall} shows the proposal and recall numbers but for the Gymnastics dataset. While TSN is more capable than the self-supervised models at higher proposal counts, it does worse than both CorrFlow DFC and TimeCycle DFC at the lower counts. It only does the best for $500$ proposals, albeit ResNet, another supervised model, has the highest score at $1000$ proposals. This result suggests that the self-supervised representations can be just as strong as the supervised representations for transfer learning. We conjecture that the discrepancy between this result and the prior one on Thumos14 is due to there being no good substitute for training on a similar distribution to the test task.

Changing our focus to the differences within the set of self-supervised models, we see from Fig.~\ref{fig:tiou-graphs} that the AMDIM model, even with the help of extra nonlinearities, can barely match CorrFlow without that aid. This suggests that AMDIM is a poor model for this task relative to CorrFlow or TimeCycle, which is contrary to how the community has ranked these models to date. This is because the literature has relied on linear probes to evaluate the transfer quality of the representations. We see here that that is an ineffective approach because there is a discrepancy in the ranking of the models in the CIFAR task and the rankings in the temporal localization tasks.

\section{Discussion}

% \cinjon{Revisit:

% 1. the random model working thing suggests that the linear probe task is not the right one. it could be that those representations *are* actually good but just need a nonlinear flip. arguably, a better approach is to always pose it as a few-shot learning task because then the quality of the representations matters a lot more and this arbitrary linear probe doesn't matter as much.
% }

% \begin{itemize}
%     \item Attention/Transformer over ensemble of available representations. 
%     \item Towards few-shot learning as a strong diagnostic for self-supervised learning. 
% \end{itemize}

We have argued in this paper that the status quo methodology for evaluating self-supervised representations is insufficient for properly adjudicating progress. Our results suggest three different model rankings. They are, in order:

CIFAR classification: ResNet,  AMDIM, Random TimeCycle, TimeCycle, and finally CorrFlow.

Thumos14 localization: TSN, then all other models.

Gymnastics localization: CorrFlow DFC, TimeCycle DFC, TSN/ResNet DFC, AMDIM DFC, CorrFlow NFC, ResNet NFC.

From this, we make four summarizing observations:
\begin{enumerate}
    \item In self-supervised learning, the bias towards success stemming from the architecture is high. We see this most clearly in the classification task with TimeCycle.
    \item The self-supervised representations are arguably just as good as the supervised ones at transferring to a hard task on a new distribution. We see this in the Gymnastics results.
    \item The accepted ordering in the self-supervised literature is not conclusive, but nor is it here. Sometimes AMDIM is better than TimeCycle and vice versa.
    \item There is no substitute for training on a similar distribution to which you test. The Gymnastics dataset helped us disambiguate this aspect of TSN and Thumos14.
\end{enumerate}

We conclude with a remark towards future research. Arguably a better approach for the community is to pose evaluation of these representations not in terms of linear or nonlinear tasks, but instead as a few-shot learning problem. Echoing our claim in the introduction, self-supervised research is driving after representation spaces that accommodate transfer learning. Instead of measuring that with the capacity of that transfer, which as we have shown is fraught with error, a better direction would be to measure it with the necessary sample complexity to yield a successful transfer.

\section{Acknowledgements}
Above all, we would like to thank Ambert Yeung and Sho Nakamori for making this research even possible. We would also like to thank Curran Phillips for his efforts in helping build the dataset, Kyunghyun Cho for his valuable feedback and guidance, Will Falcon for help with AMDIM, Will Whitney for his editor's touch, and Shubho Sengupta for being so patient. Special thanks to Adam Roberts, Doug Eck, Mohammad Norouzi, Andrew Dai, and Jesse Engel.

% \clearpage
{\small
\bibliographystyle{ieee}
\bibliography{main}
}

\end{document}